\pdfoutput=1
\documentclass[letterpaper, 10 pt, conference]{ieeeconf}  % Comment this line out if you need a4paper

\IEEEoverridecommandlockouts                              % This command is only needed if 
\usepackage{graphics} % for pdf, bitmapped graphics files
\usepackage{epsfig} % for postscript graphics files
\usepackage{amsmath} % assumes amsmath package installed
\usepackage{multirow}
\usepackage{booktabs}
\usepackage{subcaption}
\usepackage{url}
\usepackage[]{algorithm2e}

\usepackage{amsthm} {
	\theoremstyle{plain}
	\newtheorem{problem}{Problem}
}

\usepackage{color}

\title{\LARGE \bf
Ground Texture Based Localization Using Compact Binary Descriptors
}

\author{Jan Fabian Schmid$^{1,2}$, Stephan F. Simon$^{1}$, Rudolf Mester$^{2,3}$% <-this % stops a space
\thanks{$^{1}$Robert Bosch GmbH, Hildesheim, Germany \newline {\tt\small SchmidJanFabian@gmail.com}}
\thanks{$^{2}$VSI Lab, CS Dept., Goethe University, Frankfurt am Main, Germany} %\newline {\tt\small 
\thanks{$^{3}$Norwegian Open AI Lab, CS Dept., NTNU Trondheim, Norway} %\newline {\tt\small SchmidJanFabian@gmail.com}, \newline {\tt\small Rudolf.Mester@ntnu.no}}
}

\begin{document}

%This paper has been accepted in 2020 IEEE International Conference on Robotics and Automation (ICRA)
%\bigskip
%
%\textcopyright 2020 IEEE. Personal use of this material is permitted. Permission from IEEE must be obtained for all 
%other uses, in any current or future media, including reprinting/republishing this material for advertising or 
%promotional purposes, creating new collective works, for resale or redistribution to servers or lists, or reuse of any 
%copyrighted component of this work in other works.
%
%\newpage

\maketitle
\thispagestyle{empty}
\pagestyle{empty}

%%%%%%%%%%%%%%%%%%%%%%%%%%%%%%%%%%%%%%%%%%%%%%%%%%%%%%%%%%%%%%%%%%%%%%%%%%%%%%%%

% !TeX spellcheck = en_US
\begin{abstract}
Ground texture based localization is a promising approach to achieve high-accuracy positioning of vehicles.
We present a self-contained method that can be used for global localization as well as for subsequent local localization updates,
i.e. it allows a robot to localize without any knowledge of its current whereabouts,
but it can also take advantage of a prior pose estimate to reduce computation time significantly.
Our method is based on a novel matching strategy, which we call identity matching,
that is based on compact binary feature descriptors.
Identity matching treats pairs of features as matches only if their descriptors are identical.
While other methods for global localization are faster to compute,
our method reaches higher localization success rates,
and can switch to local localization after the initial localization.
\end{abstract}

%%%%%%%%%%%%%%%%%%%%%%%%%%%%%%%%%%%%%%%%%%%%%%%%%%%%%%%%%%%%%%%%%%%%%%%%%%%%%%%%

\section{INTRODUCTION}
High-accuracy localization capabilities are a precondition to enable
fully autonomous agents for tasks like freight and passenger transport.
A promising approach to this task is ground texture based visual localization using a downward-facing camera.
In contrast to approaches with a forward-facing camera,
it does not suffer from occlusion of the surrounding,
works in dynamic environments without static landmarks, avoids privacy issues,
and can be made independent of external lighting conditions.
Using ground texture allows to build infrastructure-free solutions that provide reliable centimeter precise
localization on the most relevant ground coverings like asphalt, concrete, and 
carpet~\cite{Zhang_High-Prec-Localization}.

Previous approaches require an initial localization estimation from an external source~\cite{Kelly_AGV, 
Fang_intelligent-vehicles2, Nagai_Path_Tracking, Kozak_Ranger},
making them unsuitable for a self-contained localization system,
or they are slow to compute for incremental localization updates~\cite{Zhang_High-Prec-Localization, chenstreetmap},
which limits the achievable localization accuracy.
For example, if a warehouse robot with a typical velocity of $10$\,km/h has a localization latency of $200$\,ms,
the robot moves more than $0.5$\,m during a localization update.
The path taken during the localization update can only be approximated,
which also requires additional computational effort.

We present an adaptation to the approach of Zhang et al.~\cite{Zhang_High-Prec-Localization} that performs fast
localization updates as it is able to focus on a restricted area of the map according to a prior pose estimate.
Our method employs compact LATCH~\cite{Levi_LATCH} descriptors with less than two bytes per descriptor.
Also, we introduce \emph{identity feature matching}, where only identical descriptors are considered as matches,
and use it as substitution of approximate nearest neighbor search.
These changes allow us to scale the computational effort of localization according to the confidence in a prior pose estimate,
while increasing the localization success rate compared to global methods that do not take advantage of such a prior.
Furthermore, this paper contributes the first quantitative evaluation of ground texture based localization approaches.
We compare our approach to Micro-GPS~\cite{Zhang_High-Prec-Localization}, a global method,
Ranger~\cite{Kozak_Ranger}, a local method, and StreetMap~\cite{chenstreetmap}, which can be used for both tasks.

\begin{figure}
	\centering
	\includegraphics[width=0.23\columnwidth]{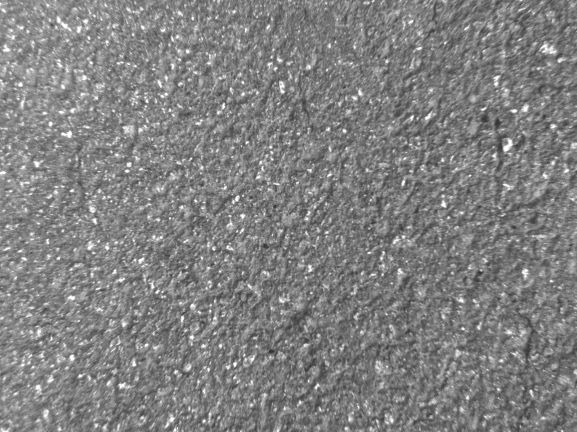}
	\includegraphics[width=0.23\columnwidth]{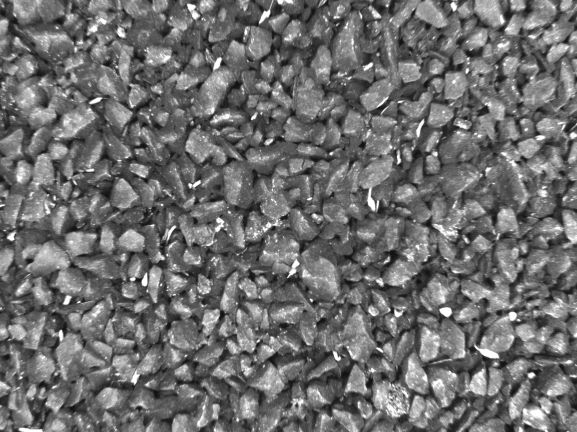}
	\includegraphics[width=0.23\columnwidth]{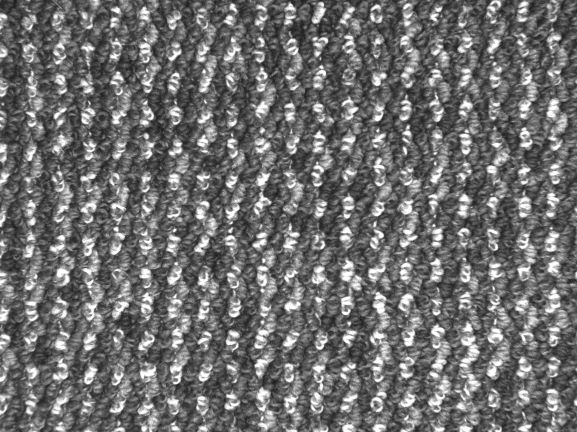}\\
	\vspace{0.1cm}
	\includegraphics[width=0.23\columnwidth]{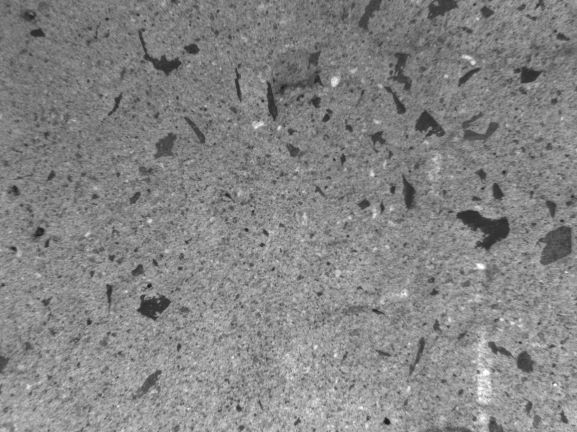}
	\includegraphics[width=0.23\columnwidth]{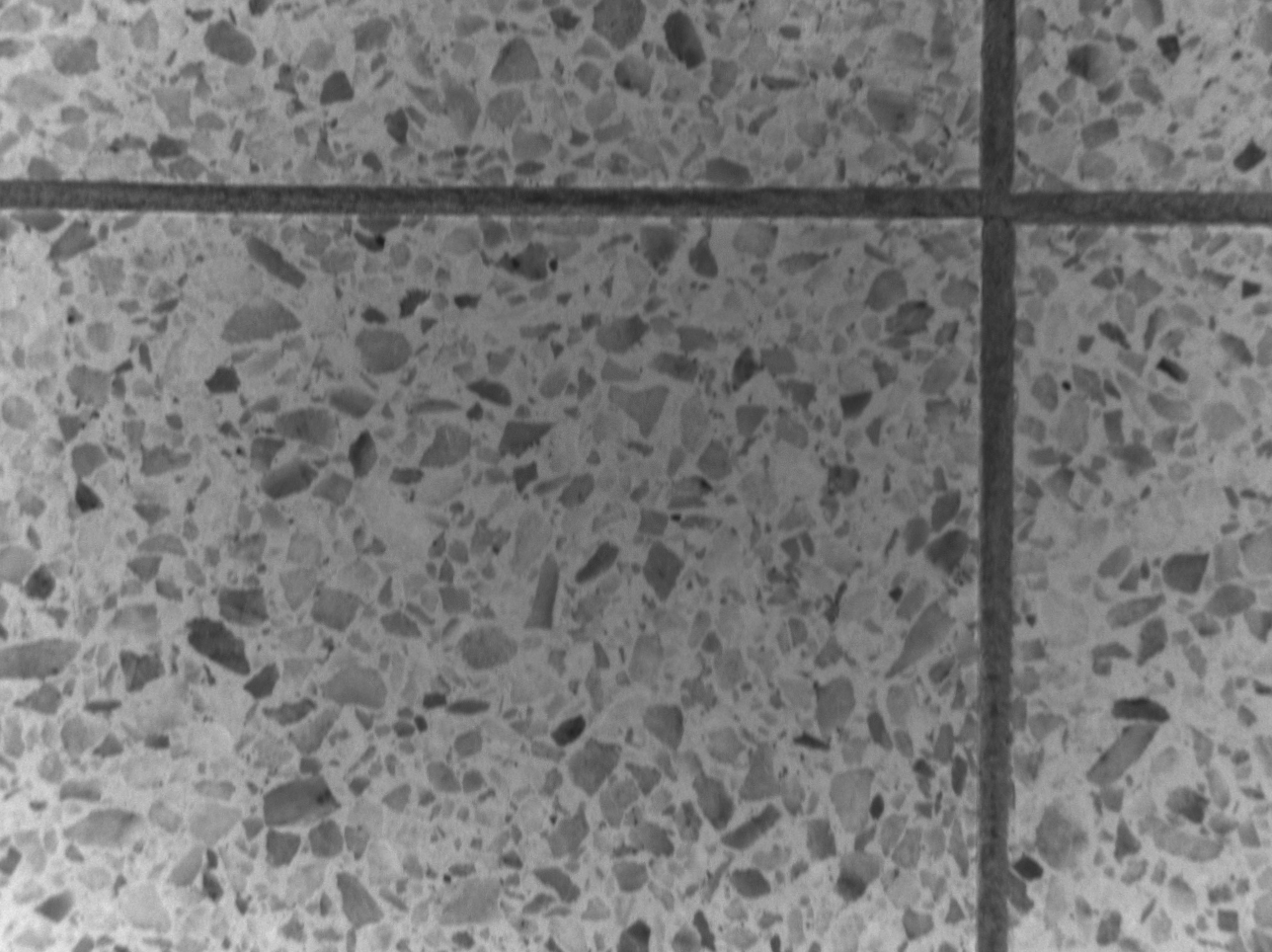}
	\includegraphics[width=0.23\columnwidth]{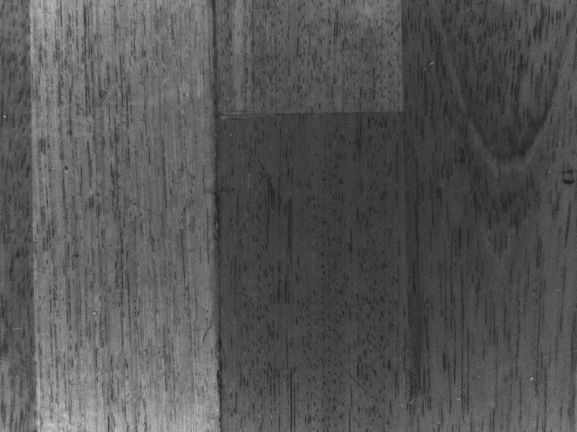}
	\caption{Examples of the ground texture image database of Zhang et al.~\cite{Zhang_High-Prec-Localization}: fine asphalt, coarse asphalt, carpet, concrete, tiles, and wood.}
	\label{fig:example_images}
\end{figure}

\section{PROBLEM STATEMENT}
\label{sec:problem_statement}
Consider an agent such as an autonomous robot with restricted operation area, e.g. a warehouse robot, 
equipped with a downward-facing camera.
To be able to take on tasks and navigate in the area,
the robot needs a map and needs to be able to localize within that map.

During the mapping phase, the agent explores the environment.
The agent gathers observations in form of ground images, and estimates their corresponding poses in the world.
We assume that these pose estimates are optimized for global consistency.
The pose estimates can be described as standard Euclidean transformations of rotation and translation in two dimensions
if we assume to have a vertically oriented pinhole camera with constant distance to a flat ground.

\begin{problem}[Mapping]
	\label{problem:mapping}
	Given a set of observations of the environment in form of ground images $I$ (the reference images) and corresponding pose estimates $T$,
	process the images to extract relevant information using an image processing function $f_m$.
	Subsequently, construct a map $M$ that stores the extracted information efficiently using a mapping function $m$ and the pose estimates $T$:
	\begin{align}
	M &= m(f_m(I), T).
	\end{align}
\end{problem}

Once a map is available, it can be used for localization.
For this purpose, the agent searches the map for visual features that correspond to the features it is currently observing.
It can be differentiated between global\,/\,initial localization without an estimation of the current pose 
($p=\emptyset$) 
and local localization with available pose estimation prior ($p\ne \emptyset$).
A localization algorithm might treat the cases $p=\emptyset$ and $p\ne \emptyset$ separately or it has a common approach to both cases.

\begin{problem}[Localization]
	\label{problem:localization}
	Given a map $M$, an observation of the environment in form of a query image $i$, an image processing function $f_l$,
	and a localization prior $p$, estimate the pose $t$ of the image using a pose estimation function $g$:
	\begin{align}
	t_{\mathrm{est}} &= g(f_l(i), M, p).
	\end{align}
\end{problem}

The estimated agent pose $t_{\mathrm{est}}$ is considered correct if it is closer to the actual pose $t$ than a threshold distance $d_t$ and if the absolute angle between the two Euclidean transformations is smaller than an orientation threshold $o_t$.

\section{RELATED WORK}
\label{sec:related_work}
We consider map-based localization approaches for robots equipped with a downward-facing camera.
These are methods that allow to perform absolute localization within a confined area.
Methods for incremental localization, that estimate the vehicle pose relative to a previous pose, accumulate drift
and therefore need to be accompanied by an error correction mechanism, e.g. an absolute localization method.
Examination of these methods is out of our scope.

Global methods do not require a localization prior \cite{Zhang_High-Prec-Localization, chenstreetmap},
while local methods do \cite{Kelly_AGV, Fang_intelligent-vehicles2, Nagai_Path_Tracking, Kozak_Ranger},
and therefore have to be initialized in another way, e.g. using GPS.

In order to localize, it is necessary to find correspondences between the mapped reference images and the current view of the autonomous agent.
This can be done with photometric approaches~\cite{Kelly_AGV},
that compare images based on a function of image intensity values, e.g. normalized cross correlation,
or with feature-based approaches,
that propose well-matching features as correspondences \cite{Zhang_High-Prec-Localization, Fang_intelligent-vehicles2, Nagai_Path_Tracking, Kozak_Ranger, chenstreetmap}.
These features are representations of characteristic image regions~\cite{Pratt_Image-Processing}.
The position of a feature in the image is specified by its \emph{keypoint}.
Additionally, size and orientation might further specify the feature.
The \emph{feature descriptor} describes the local environment of a keypoint.
For keypoints it is important that they are repeatable,
i.e. the same keypoints for the same physical locations are found for varying conditions of recording like camera position and illumination.
Descriptors should take similar values for corresponding keypoints,
and distinctively different values for non-corresponding ones.

\subsection{Global localization approaches}
Micro-GPS is a localization pipeline proposed by Zhang et al.~\cite{Zhang_High-Prec-Localization}.
They rely on SIFT for feature extraction,
and construct an efficient approximate nearest neighbor (ANN) search structure for feature matching,
exploiting the fact that the scale of corresponding features remains essentially constant for images of a downward-facing camera with stable height.
Per reference image $50$ randomly sampled features are inserted into the search structure.
During localization, all query image features are used for feature matching.
Each feature from the query image is paired with its ANN.
A voting procedure is employed for outlier rejection.
For this purpose, a voting map is created, which is a grid with a size that corresponds to that of the mapped environment.
Then, each retrieved ANN votes for the grid cell containing the camera position that would explain the occurrence of the proposed correspondence.
Finally, only the correspondences voting for the grid cell that obtained the most votes are used to estimate the camera pose with a RANSAC procedure.

Chen et al. developed StreetMap~\cite{chenstreetmap}, which is able to make use of a localization prior,
but does not require one.
While there is also a version specifically for tiled ground textures,
we only consider the feature-based variant.
If no prior is available,
they use bag of words (BoW) image retrieval \cite{Lopez_BagOfBinaryWords} to find similar reference images to the query image.
For this purpose, BoW representations of the images are computed using the SURF~\cite{Bay_SURF} feature descriptors extracted from them.
After retrieval of the most similar reference images,
their features are matched to the features of the query image.
For each feature of the query image,
they search for its nearest neighbor from the reference images and subsequently filter these matches with the \emph{ratio test constraint} \cite{Lowe_SIFT2},
which requires that the most similar reference descriptor is significantly closer to the query descriptor than the 
second most similar one. % TODO could be removed
The Euclidean transform of the camera pose is finally estimated using RANSAC.

\subsection{Local localization approaches}
Kelly et al. \cite{Kelly_AGV, Kelly_Mosaics} developed a photometric localization approach using normalized cross correlation for template matching to find corresponding image patches between query and reference images.
They construct a ground map of statistically normalized pixel intensity values. % using image stitching.
During localization, the output of a Kalman filter is used as a localization prior.
Peaks of a texture score function, which depends on the local intensity gradient of the pixels,
are used to define up to $16$ image patches for template matching.
The difference between predicted and observed positions of these image patches is combined into a pose update.

The localization pipeline of Fang et al.~\cite{Fang_intelligent-vehicles2, Fang_intelligent-vehicles1} relies on the iterative closest point (ICP) algorithm to align reference images during mapping and to register query images for localization.
The point clouds needed for this purpose are built using extracted corner and edge features. % from the images.
For the final pose estimation the results of a robust ICP variant are fused with odometry information in an extended Kalman filter.

Nagai and Watanabe~\cite{Nagai_Path_Tracking, nagai2007mobile} propose a method that avoids the need for a globally consistent map.
Instead, they construct a sparse spatial map of images.
Whenever the autonomous system approaches a reference image stored in the map,
correspondences between query and reference image are used to correct for the drift that accumulated since the last 
absolute localization step.
Image transformations are estimated through minimization of the reprojection error,
which is measured as cross-correlation of intensity values. 

Kozak and Alban~\cite{Kozak_Ranger} developed Ranger,
a method that enables localization at high vehicle speeds of up to $120$\,km/h.
Ranger extracts ORB~\cite{Rublee_ORB} feature descriptors at CenSurE~\cite{Agrawal_CenSurE} keypoints.
Based on the current prior, the closest reference image is selected.
Then, correspondences are generated through nearest neighbor matching of features from this reference image and the query image.
A \emph{cross check} is performed to reject incorrect matches.
This means that in order for the reference image feature $F_r$ and the query image feature $F_q$ to be considered a match,
among all reference features $F_r$ needs to be the nearest neighbor of $F_q$ and
among all query features $F_q$ needs to be the nearest neighbor of $F_r$.
If afterwards at least $25$ correspondences remain, these are used to estimate the camera transformation with RANSAC.
Otherwise, the procedure is repeated with the next closest reference image (or localization is aborted due to timeout).

As mentioned, StreetMap is able to make use of a prior as well \cite{chenstreetmap}.
Instead of selecting reference images based on BoW similarity,
the images with shortest spatial distance to the prior are taken into consideration.
 
\section{METHOD}
\label{sec:method}
We adapt Micro-GPS, the localization pipeline of Zhang et al. \cite{Zhang_High-Prec-Localization}.
Micro-GPS achieves reliable high-precision localization on most of the evaluated ground textures,
but it requires more than hundred milliseconds for each localization request,
even on a fast computer with hardware acceleration.

We identify the construction of a global approximate nearest neighbor (ANN) search structure for feature matching, as a major drawback of Micro-GPS.
It allows to perform efficient feature matching between query and reference images;
however, the structure represents a fixed set of reference images and needs to be recomputed whenever another image is added to the map.
Updating a reference image with a more recent recording requires recomputation as well.
Also, using this matching technique means that correspondences are always searched globally.
The method cannot use a localization prior to reduce the number of considered reference images.

We tackle these drawbacks, using \emph{identity matching} in conjunction with compact binary descriptors.

For feature extraction, we determine keypoints and their orientations using SIFT~\cite{Lowe_SIFT2},
and compute feature descriptors with LATCH~\cite{Levi_LATCH}.
The SIFT feature detector locates regions of interest as local extrema on a Gaussian scale-space pyramid.
LATCH computes binary descriptors for keypoints through the comparison of image patch triplets.
An anchor patch $p_a$, is extracted at the position of the keypoint
and is then compared to two surrounding image patches ${p_1, p_2}$.
Each bit value of the LATCH descriptor is evaluated by one triplet,
each of which specifies a unique placement of $p_1$ and $p_2$.
A triplet is evaluated to $1$ if $p_a$ is more similar to $p_1$ than to $p_2$ and to $0$ otherwise.
We take advantage of the original LATCH triplet arrangements, which have been optimized by the authors.
The order of the employed triplets is a ranking based on how many times a triplet has the same value for 
corresponding keypoints and different values for non-corresponding ones.
Furthermore, strongly correlating triplets were removed.
In our case, we use only the first $15$ triplets as compact binary descriptors,
which results in the highest success rate for our number of extracted features ($850$).
A higher number of bits increases the inlier-to-outlier ratio,
but decreases the absolute number of inliers.
To compensate for this, we would have to extract more features,
increasing computation cost and memory consumption.

Our matching strategy proposes only those pairs of descriptors as matches that have identical values.
Identity matching can be implemented efficiently as table lookup,
i.e. row $i$ of the table contains references to descriptors whose decimal representation of their binary string is 
equal to $i$.
For feature matching of binary descriptors with a dimensionality of $n$ ($n=15$ in our case),
a table of length $2^n$ is created and filled with pointers to the reference features.
Then, to find matches for a feature,
it is sufficient to retrieve the pointers of the table row that corresponds to the feature descriptor.

In contrast to the ANN search index employed by Zhang et al.,
it is not necessary to compute one search structure for the entire map,
but feature matching can be performed on an image to image basis.
Accordingly, during mapping, we create a descriptor table for each reference image.
If a localization prior is available,
only the tables of the closest reference images are considered for feature matching.
For global localization without prior, all tables are considered.

The use of identity matching with compact binary descriptors leads to a large number of incorrectly proposed matches (outliers).
E.g. for global localization, typically less than $0.015\%$ of matches can be considered correct correspondences (inliers).
This is why we employ the voting procedure of Micro-GPS~\cite{Zhang_High-Prec-Localization} for outlier rejection.
Here, the outlier matches distribute their votes for the current camera position equally on the voting map,
while the inlier votes are concentrated in a narrow region (see Figure \ref{fig:voting_map}).
Subsequently, the matches that voted for the map cell with most votes are used for a RANSAC-based estimation of the camera pose.

\section{EVALUATION}
\label{sec:evaluation}
For evaluation, we use the ground texture image database of Zhang et al.~\cite{Zhang_High-Prec-Localization}.
It contains datasets of six texture types (see Figure \ref{fig:example_images}), recorded with a Point Grey camera.
The 8-bit gray scale images have a resolution of $1288$ by $964$ pixels,
covering an area of about $0.2\mathrm{m}\times 0.15\mathrm{m}$.
For each texture type, Zhang et al. provide a set of reference images for mapping,
consisting of about $2000$ to $4000$ partially overlapping recordings.
Furthermore, the database provides sequences of ground images that were recorded on paths independent to the paths taken for mapping,
which we use to evaluate localization performance.

Prior to the evaluation, we find the best suited parameters of the examined methods,
if not specified by the respective authors, on a training set of $100$ query images per ground texture type.
Subsequently, a separate set of $500$ images per texture type is used to evaluate our experimental setups.

The employed hardware consists of a E3-1270 Intel Xeon CPU at $3.8$\,GHz,
and a Quadro P2000 Nvidia graphics card (used to compute SIFT features in Micro-GPS).

We separately evaluate localization methods for initial localization without available prior and
for subsequent local localization with available prior.
Our main performance metric is the pose estimation success rate,
i.e. the proportion of localization queries for which the estimated pose $t_{\mathrm{est}}$ is closer to the actual pose $t$ than $d_t$ with an absolute angle difference of less than $o_t$.
We adopt the thresholds of Zhang et al. \cite{Zhang_High-Prec-Localization} of $d_t=30\text{\,pixels}$ ($4.8\text{\,mm})$ and $o_t=1.5 \text{\,degrees}$.

\subsection{Evaluation of global localization}
Besides our method, we evaluate Micro-GPS~\cite{Zhang_High-Prec-Localization},
for which the code is provided by the authors,
and StreetMap~\cite{chenstreetmap}, which is re-implemented according to the paper.

\subsection{Evaluation of local localization}
For our examination of localization performance with available localization prior,
we evaluate our method, StreetMap, and Ranger \cite{Kozak_Ranger},
which we re-implemented according to the system description of the authors.

Here, we evaluate pose estimation success rates for varying accuracies of the localization prior.
The prior is created by taking the ground truth camera position of the query image and by shifting it into a randomly sampled direction.

All of the evaluated local methods use the prior only to select a subset of closest reference images to the current 
pose estimate.
In our experiments, we choose the number of considered images in respect to the available prior accuracy,
based on empiric evaluation of the number of closest images to the prior position
that is necessary to ensure that the closest images to the actual camera position are included.

\section{IMPLEMENTATION}
\label{implementation}
In the following, we present implementation details of the evaluated localization methods.
We describe the image processing, mapping, and localization functions.
For all of the evaluated methods,
feature extraction is the same for reference and query images ($f_m$ is similar to $f_l$).

\subsection{Our method}
\textbf{Image processing:}
We employ the SIFT implementation of the OpenCV 4.0 library~\cite{opencv_library} to extract keypoints.
The number of layers per pyramid octave is set to $11$, the contrast threshold to $0.005$,
the edge threshold to $13$, and the sigma of the employed Gaussian filter is set to $8.5$.
Only the $850$ keypoints with largest response values are kept.
Then, we extract for each keypoint the first $15$ bit of the OpenCV LATCH descriptor.
In order to deal with varying image orientations,
we use the LATCH variant that rotates the considered image patch according to the keypoint orientation.
The half-size of the evaluated patches is set to $8$, and the sigma of the employed Gaussian smoothing is set to $2.2$.

\textbf{Mapping:}
For each reference image, the identity matching table is built.
These tables are sparsely populated,
which is why we implement them as dictionaries that map descriptor values to lists of indexes from features with that
descriptor value.
To use available priors,
a k-dimensional tree (k-d tree) is constructed from the pose estimates of the reference images,
using the nanoflann library~\cite{nanoflann}.

\textbf{Localization:}
If a localization prior is available,
only the closest reference images are considered.
Otherwise, we perform identity matching with all reference images.
The retrieved matches are used to cast votes for the corresponding camera positions on a voting map.
The cell size of the voting map grid is set to $75 \times 75$ pixels ($12 \times 12$\,mm).
We select the matches that voted for the voting map cell with most votes
and perform RANSAC based pose estimation with them.

\subsection{Micro-GPS}
\textbf{Image processing:}
Zhang et al.~\cite{Zhang_High-Prec-Localization} use SiftGPU\footnote{\url{https://github.com/pitzer/SiftGPU}} to extract SIFT features.
As for all other evaluated localization methods, features are extracted from full-scale images.
The authors employ principal component analysis (PCA) dimensionality reduction to reduce the size of the SIFT 
descriptors.
In our case, the PCA basis for that purpose is created using the entire set of reference images of the currently evaluated texture type.
We use 16-dimensional descriptors, which the authors found to perform better than 8-dimensional ones~\cite{Zhang_High-Prec-Localization}.

\textbf{Mapping:}
Of each reference image $50$ 16-dimensional SIFT features are randomly sampled.
The authors assume that corresponding features will have similar scale.
Therefore, they use the scale information to divide the set of reference features into $10$ groups.
For each group, they construct an ANN search index with the FLANN library~\cite{Muja_FLANN}.

\textbf{Localization:}
For each 16-dimensional SIFT feature of the query image, its ANN reference feature is retrieved,
using the search index corresponding to the feature's scale.
Each of the obtained matches casts a vote for the camera position on a voting map with a cell size of $50 \times 50$ 
pixels ($8 \times 8$\,mm).
Afterwards, the matches that voted for the voting map cell with most votes are used for RANSAC pose estimation.

\subsection{StreetMap (without prior)}
\textbf{Image processing:}
We extract SURF features using OpenCV~\cite{opencv_library}, using 4 pyramid octaves with 3 layers each,
and a Hessian threshold of $20$.
Per image the $1000$ features with largest response values are kept for further processing.

\textbf{Mapping:}
For each image, a BoW representation is computed based on the retrieved SURF features,
using the FBOW library\footnote{\url{https://github.com/rmsalinas/fbow}}.
The vocabulary for that purpose was computed beforehand,
using default parameters of the library and the extracted SURF features of $1000$ images per texture type.

\textbf{Localization:}
The number of considered reference images is reduced by $80\%$,
by selecting the most similar ones to the query image based on their BoW representations.
This value is a trade-off between localization performance and computation time.
For matching, we find for each query image feature the most similar reference feature from the remaining reference images,
using the L2 norm and the OpenCV brute force feature descriptor matcher.
A ratio test with a threshold of $0.9$ is employed for outlier rejection.
Poses are estimated in a RANSAC fashion, using the obtained feature matches.

\subsection{StreetMap (with prior)}
\textbf{Image processing:}
OpenCV~\cite{opencv_library} SURF features are extracted from an image pyramid with $5$ octaves with $4$ layers each.
The Hessian threshold for keypoint rejection is set to $20$,
and only the $768$ features with largest responses are kept.

\textbf{Mapping:}
A k-d tree~\cite{nanoflann} is built from the reference image positions.

\textbf{Localization:}
The procedure is the same as for global localization,
but the considered reference images are selected based on closeness to the prior,
using the k-d tree.

\subsection{Ranger}
\textbf{Image processing:}
Kozak and Alban~\cite{Kozak_Ranger} use CenSurE~\cite{Agrawal_CenSurE} keypoints, which are not robust to the image orientation.
For street vehicles, robustness to orientation is not required because typically the vehicle orientation is the same during mapping and localization.
In our evaluation, however, image orientations during mapping and localization are independent of each other.
Therefore, we exchange CenSurE with AKAZE~\cite{Alcantarilla_AKAZE} keypoints,
which among the OpenCV~\cite{opencv_library} keypoint detectors achieved the best results for our Ranger implementation.
The best parameters we found for AKAZE are a response threshold of $0.00001$, and a single image pyramid octave with two layers.
Up to $1250$ keypoints with largest response values are kept per image.
For feature description, we employ the rotation invariant BRIEF descriptor of OpenCV with its full size of $64$ bytes.

\textbf{Mapping:}
A k-d tree~\cite{nanoflann} is built from the reference image positions.

\textbf{Localization:}
Features of query image and the closest reference image are matched using the OpenCV brute force feature descriptor matcher with Hamming norm.
For outlier rejection, a cross check is performed.
The remaining feature matches are used for RANSAC based pose estimation.
If the estimated pose is supported by at least $25$ matches,
it is used as final output of the method.
Otherwise, the procedure of matching and pose estimation is repeated with the next closest reference image, and so on.
If the condition is not met by any of the considered reference images,
we use the pose estimation that had the most inliers.

\section{RESULTS}
\label{sec:results}
\begin{figure}[t]
	\centering
	\includegraphics[width=1.0\columnwidth]{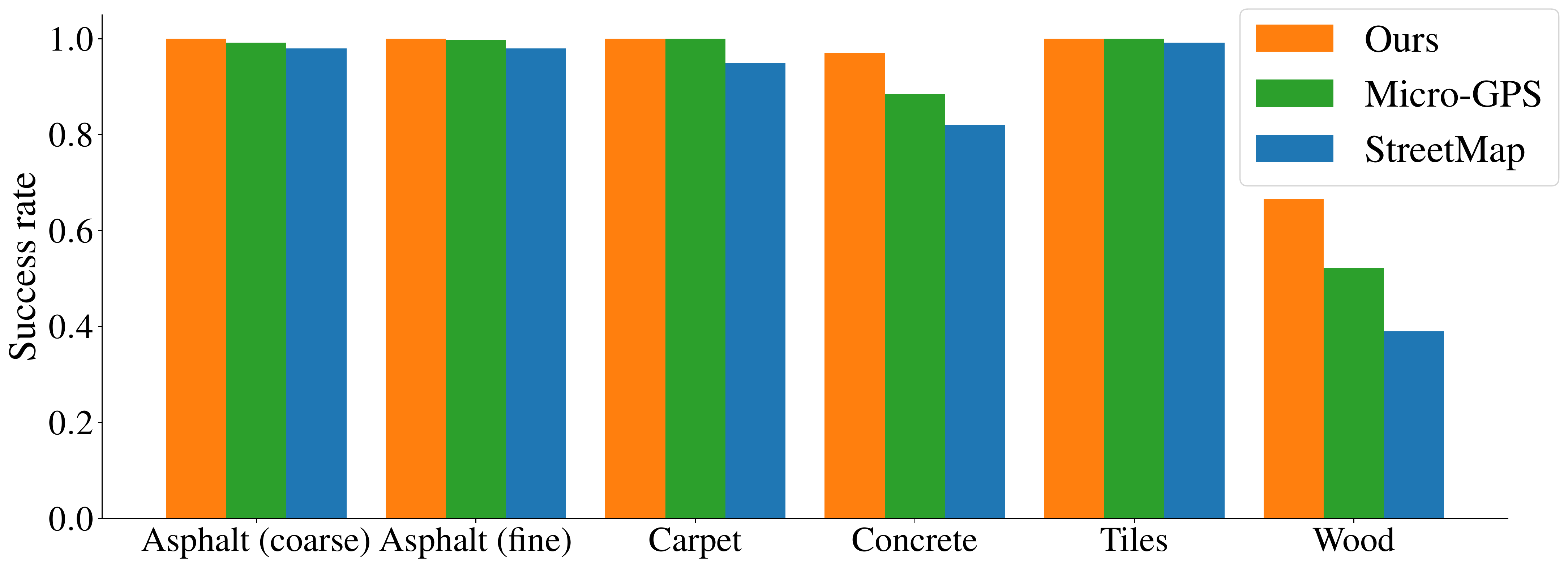}
	\caption{Pose estimation success rates for global localization.}
	\label{fig:without_prior}	
\end{figure}

Pose estimation success rates for our experimental setup for global localization are presented in Figure \ref{fig:without_prior}.
We observe that both types of asphalt, carpet, and tiles are particularly well suited for ground texture based localization,
as all three evaluated methods reach almost perfect success rates.
The situation is different for concrete and wood.
While our method is still able to localize correctly in $97.0\%$ of the test cases on concrete texture,
the original Micro-GPS reaches only $88.4\%$ success rate and StreetMap $82.0\%$.
For wooden texture, our method is again the best performing method,
but only achieves a success rate of $66.6\%$,
while Micro-GPS and StreetMap have $51.4\%$ and $39.0\%$, respectively.
Further analysis shows that lower success rates can be explained with lower numbers of inliers among the matched features.
During localization, our method identifies on average more than $40$ inliers for asphalt, carpet and tiles,
but only $31.5$ for concrete and $9.7$ for wood texture images.
One explanation for this is, that among the evaluated textures the wooden images are most similar to each other,
resulting in lower keypoint repeatability.
In fact, using pairs of synthetically transformed images,
we find that wood is the most challenging texture for keypoint detectors to retrieve corresponding keypoints \cite{Schmid_Survey}.

\begin{figure}[t]
	\centering
	\includegraphics[width=1\columnwidth]{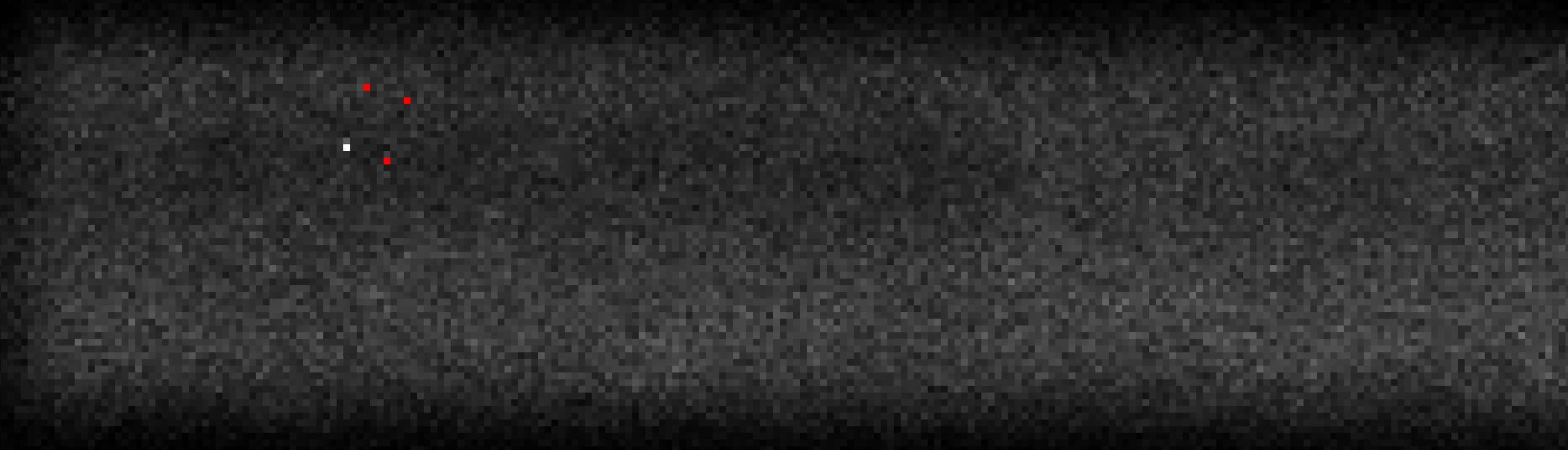}
	\caption{Cutout of a voting map from a successful global localization attempt. Brightness indicates the voting numbers. Matches vote for the position of the upper left image corner. The true positions of the other 3 corners are depicted in red.}
	\label{fig:voting_map}	
\end{figure}

A voting map is illustrated in Figure \ref{fig:voting_map}.
For better visualization, we doubled the voting cell size.
One cell, which is corresponding to the actual camera position,
received the most votes, while outlier votes are randomly distributed.

\begin{figure}[t]
	\begin{subfigure}[c]{0.493\columnwidth}
		\includegraphics[width=1\columnwidth]{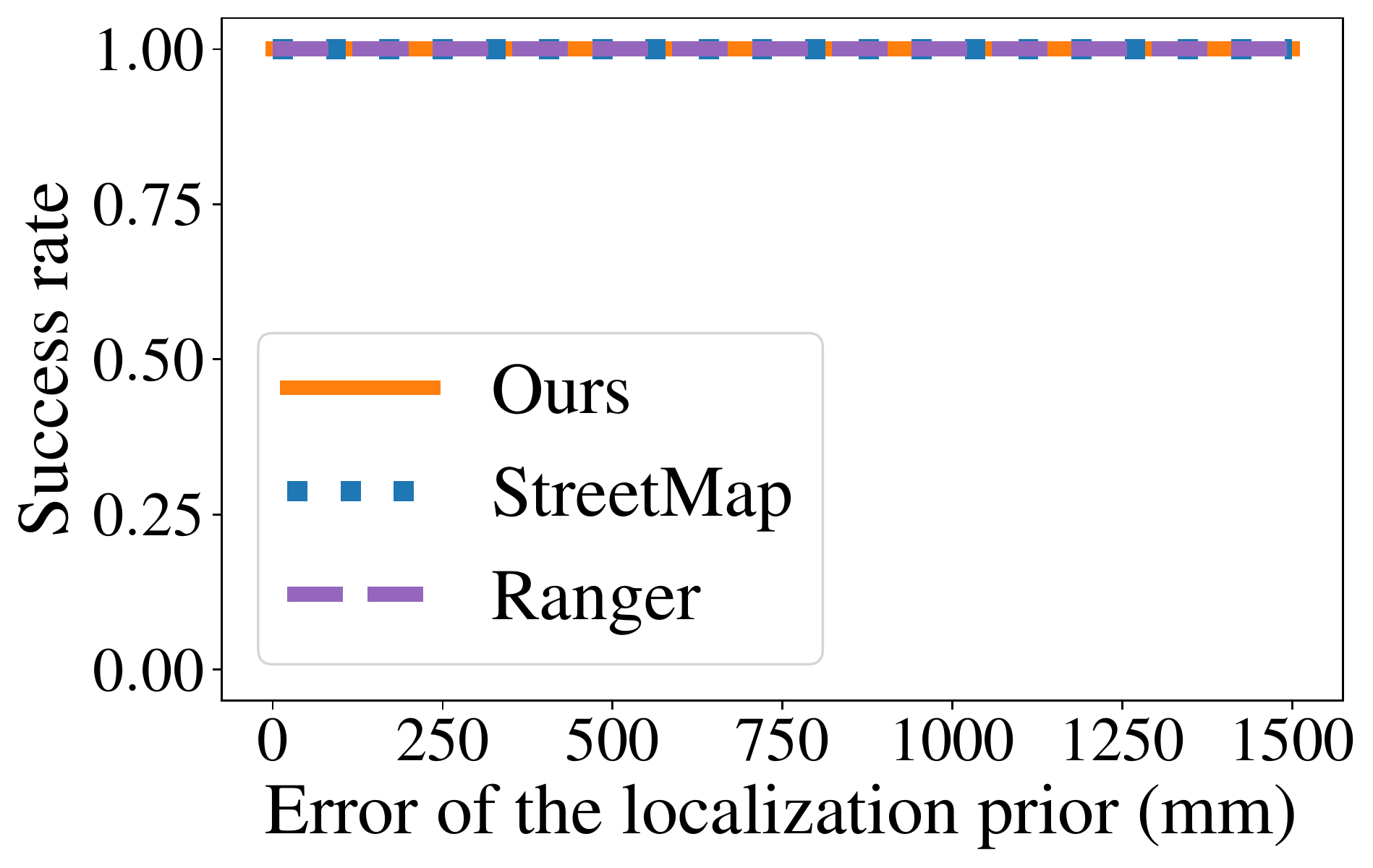}
		\subcaption{Asphalt (coarse)}
	\end{subfigure}
	\hfill
	\begin{subfigure}[c]{0.493\columnwidth}
		\includegraphics[width=1\columnwidth]{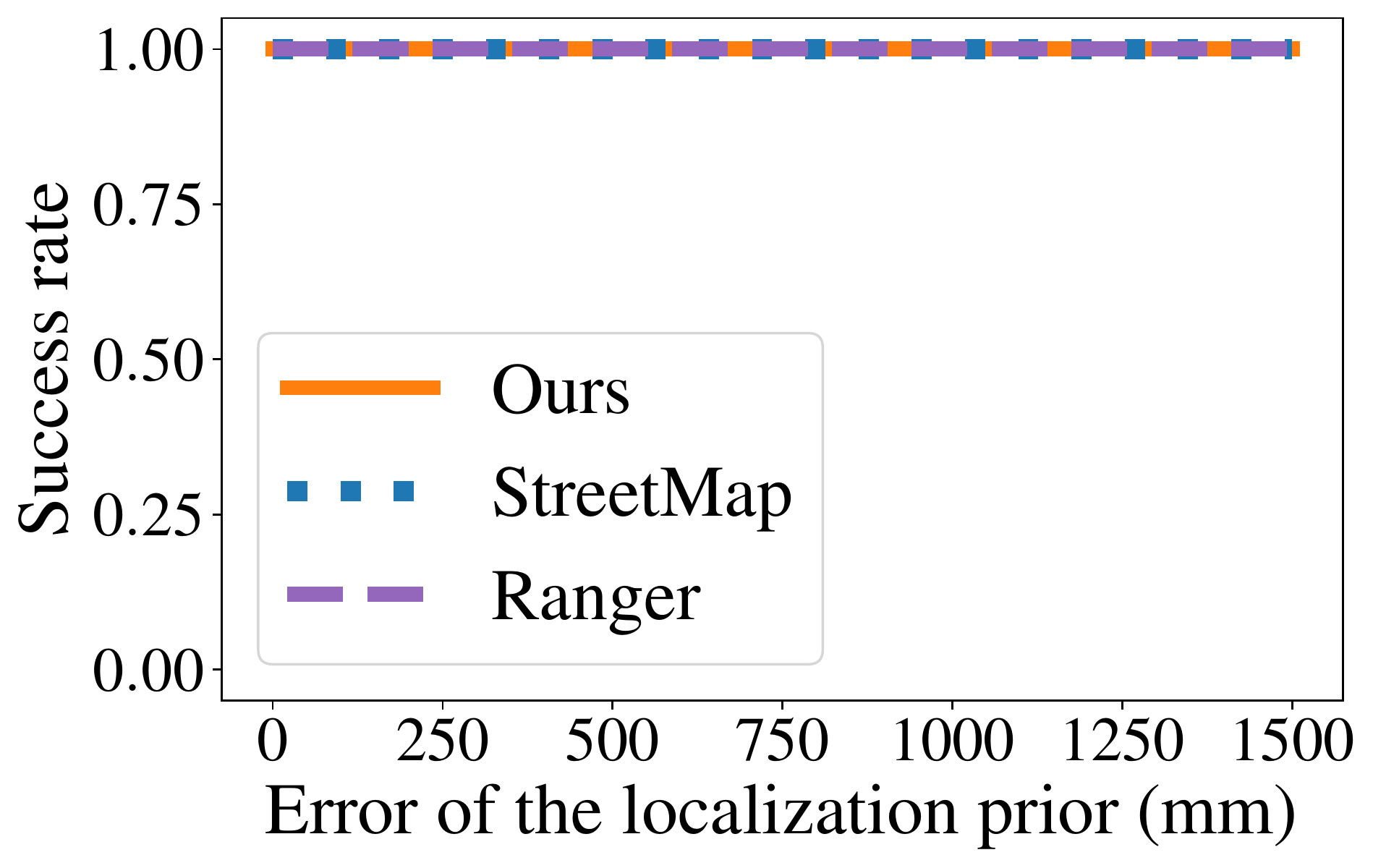}
		\subcaption{Asphalt (fine)}
	\end{subfigure}
	\begin{subfigure}[c]{0.493\columnwidth}
		\includegraphics[width=1\columnwidth]{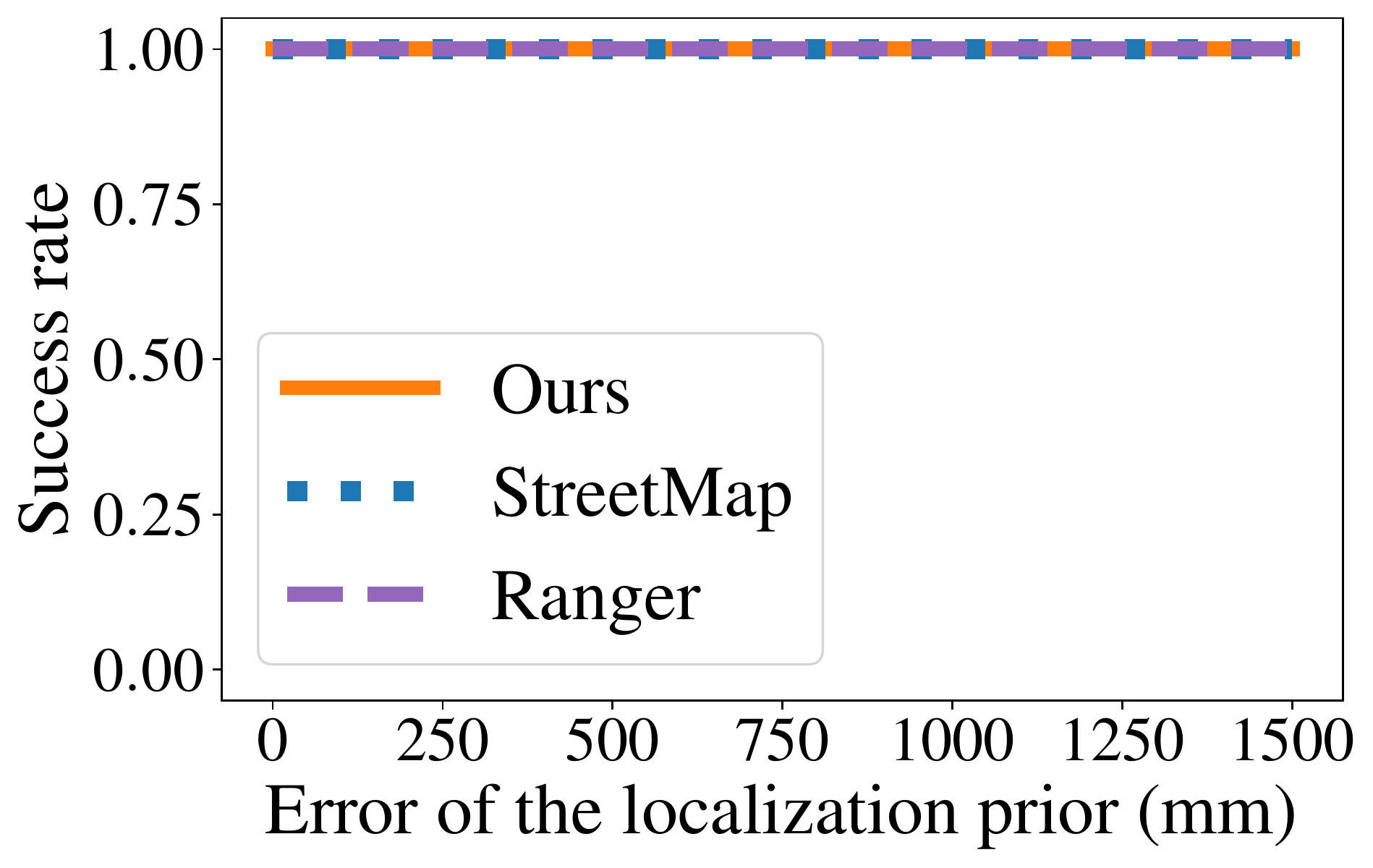}
		\subcaption{Carpet}
	\end{subfigure}
	\hfill
	\begin{subfigure}[c]{0.493\columnwidth}
		\includegraphics[width=1\columnwidth]{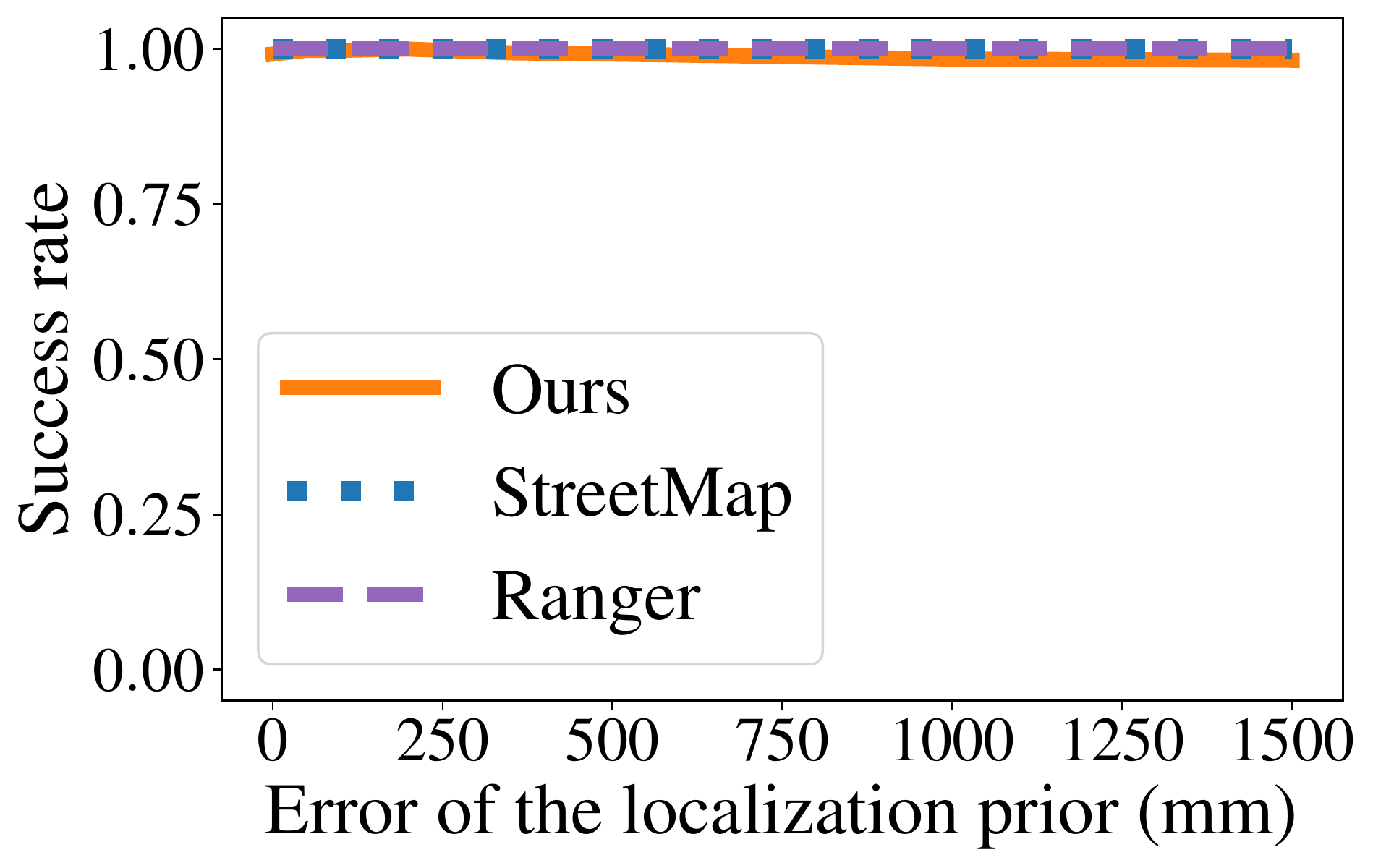}
		\subcaption{Concrete}
	\end{subfigure}
	\begin{subfigure}[c]{0.493\columnwidth}
		\includegraphics[width=1\columnwidth]{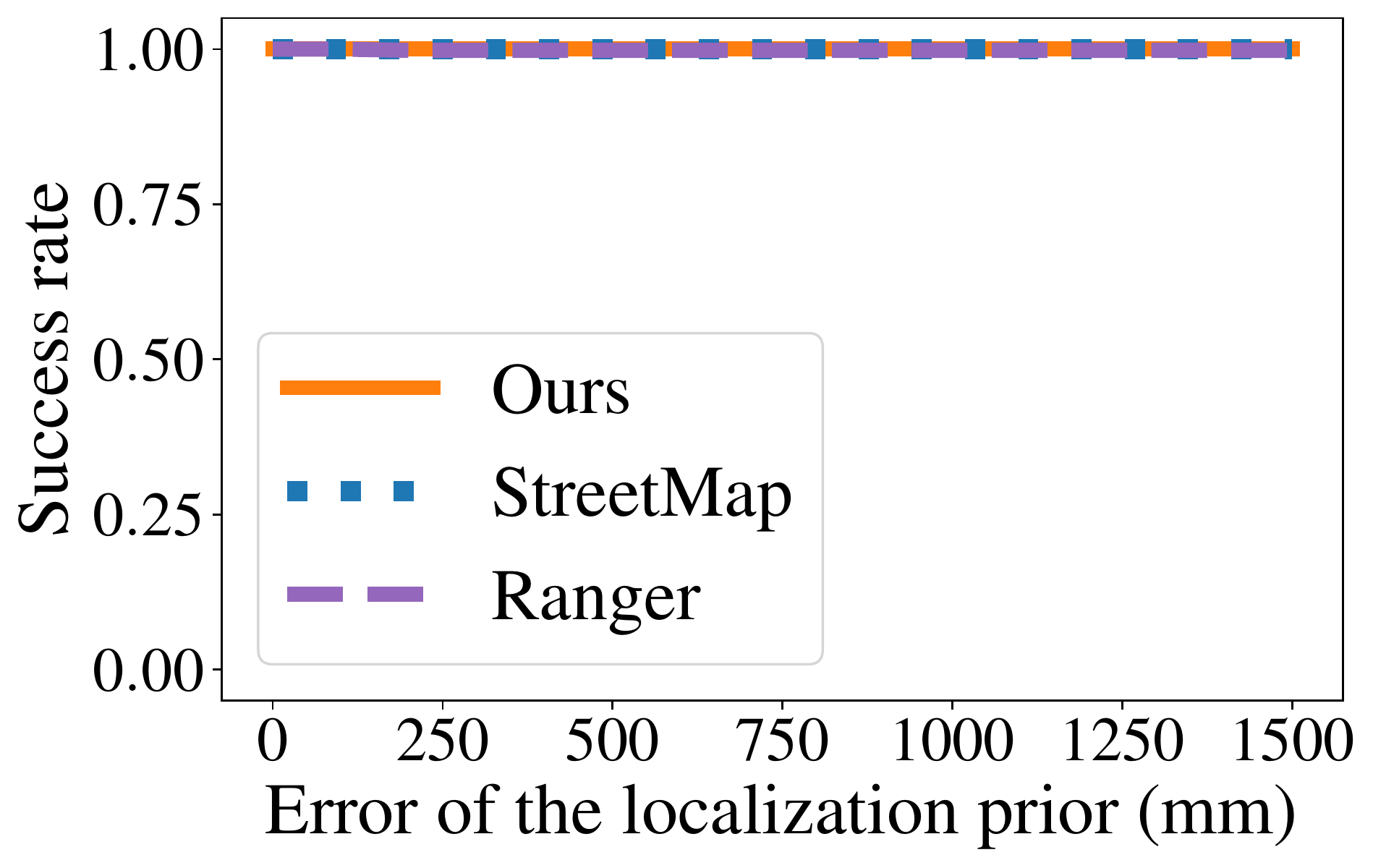}
		\subcaption{Tiles}
	\end{subfigure}
	\hfill
	\begin{subfigure}[c]{0.493\columnwidth}
		\includegraphics[width=1\columnwidth]{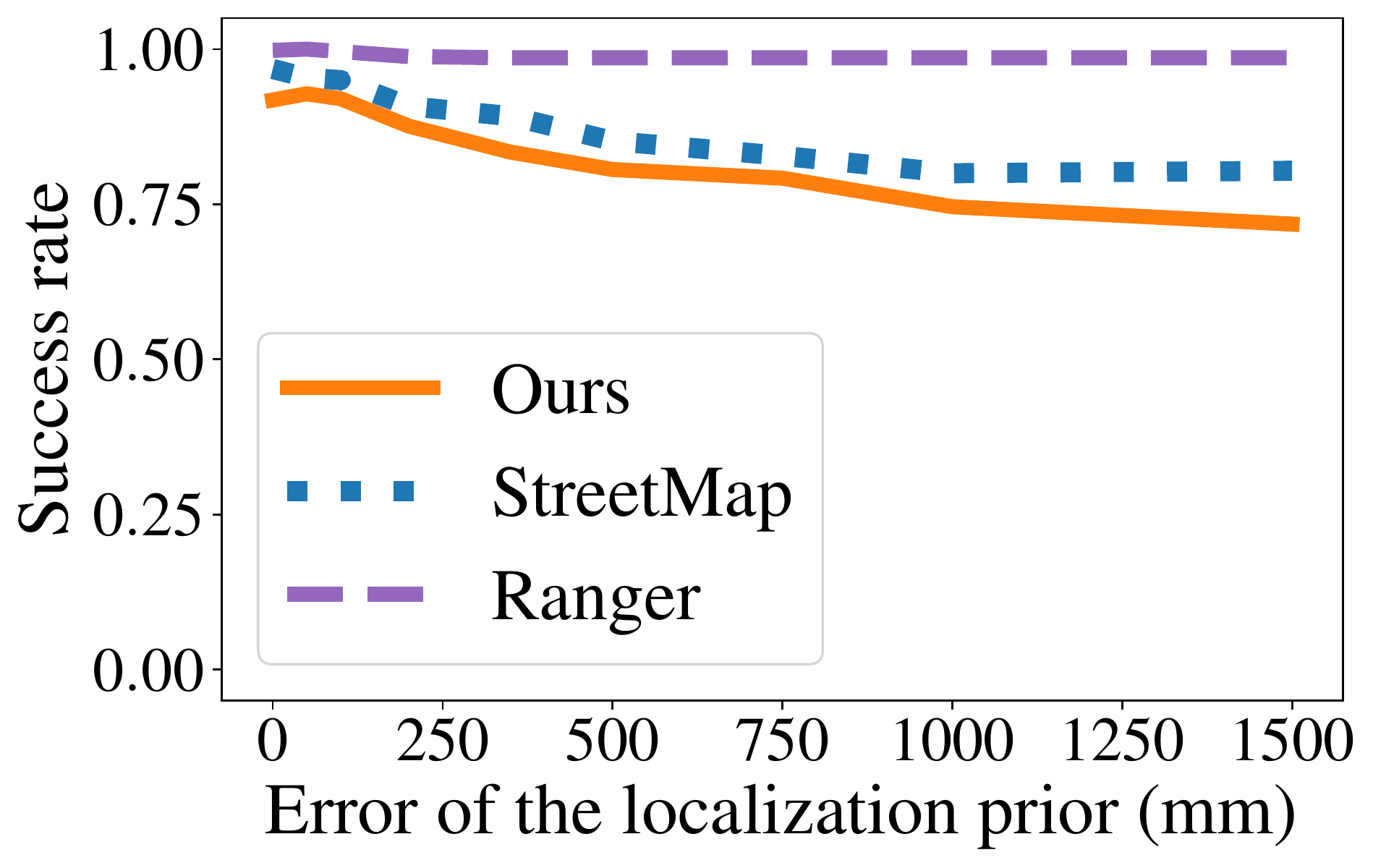}
		\subcaption{Wood}
	\end{subfigure}
	\caption{Pose estimation success rates for local localization.}
	\label{fig:with_prior}	
\end{figure}

For local localization, results are presented in Figure \ref{fig:with_prior}.
As explained previously, we empirically determined suitable numbers of reference images that are taken into consideration for a certain prior accuracy.
The corresponding fixed numbers can be found in Table \ref{table:computation_time},
they are chosen rather conservatively to avoid a situation in which localization with the available set of reference images is not possible.

On both asphalt types, carpet, concrete, and tiles,
all three evaluated methods are almost always able to localize correctly.
Again, wood (Figure \ref{fig:with_prior}(f)) presents itself as the most challenging ground texture type.
With decreasing prior accuracy, localization success rates of StreetMap and our method decline.
Again, this can be explained with a low number of inlier matches for wood,
which leads to a less significant inlier voting peak than there is for other textures.
For increasing numbers of considered reference images,
the number of outlier votes increases,
and it becomes more likely that variations in the distribution of outlier votes cause higher voting peaks than the inliers.
Similarly, the inlier-to-outlier ratio of StreetMap decreases with increasing numbers of reference images,
while Ranger considers one reference image after the other and is therefore robust to this problem.

\begin{figure}[t]
	\centering
	\includegraphics[width=0.621\columnwidth]{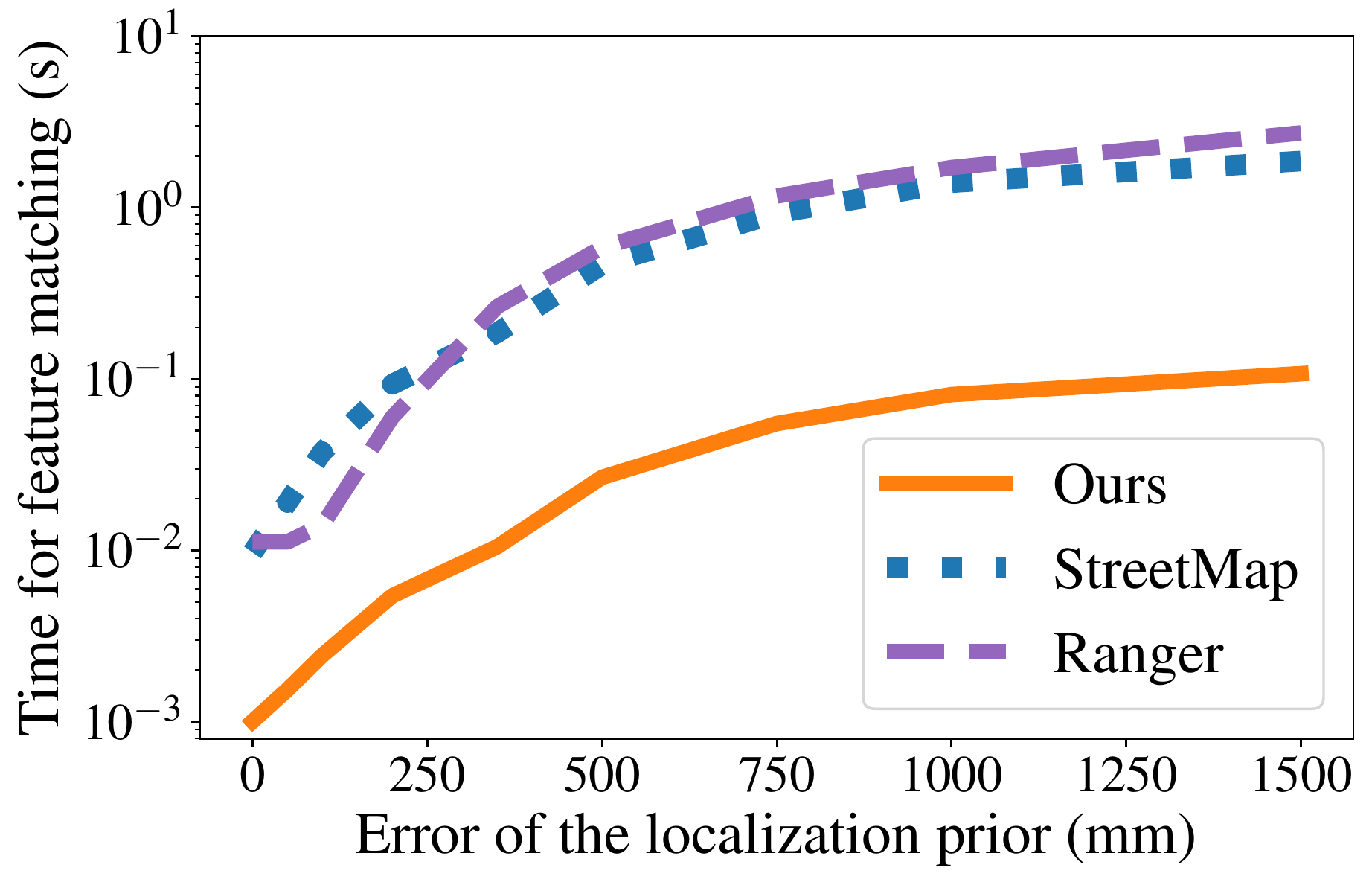}
	\caption{Required computation for feature matching on the carpet dataset for varying prior accuracies.}
	\label{fig:matching_time_with_prior_Carpet}
\end{figure}

On wood, our approach is outperformed by both StreetMap and Ranger.
However, they become slow for larger errors of the prior,
due to the use of nearest neighbor matching,
computing distances between all possible pairings of query feature descriptors and reference feature descriptors.
Figure \ref{fig:matching_time_with_prior_Carpet} presents the required computation time of feature matching for the three evaluated localization methods on the carpet test set.
Using a prior with an expected error of $0.35$\,m,
it takes $0.19$\,s to match features for StreetMap and $0.26$\,s for Ranger,
while our method takes only $0.01$\,s.
If the expected prior error is $1.5$\,m,
feature matching for StreetMap takes $1.87$\,s and $2.72$\,s for Ranger, but only $0.11$\,s for our method.

\begin{table}[t]
	\centering
	\caption{Localization time (without feature extraction) on the carpet dataset dependent on the accuracy of the 
	prior.}
	\begin{tabular}{cccc} 
		\toprule
		Error of the locali- & Number of considered & \multicolumn{2}{c}{Computation time (ms)} \\
		zation prior (mm) & reference images & Ours & Micro-GPS \\
		\midrule
		0 & 5 & 1.60 &   \\
		50 & 10 & 2.42 &  \\
		100 & 20 & 3.85 &  \\
		200 & 50 & 8.12 &   \\
		350 & 100 & 15.25 &  \\
		500 & 250 & 36.74 &   \\
		750 & 500 & 73.87 &    \\
		1000 & 750 & 108.48 &   \\
		1500 & 1000 & 143.65 &   \\
		No prior & 2014 & 286.47 & 145.55 \\
		\bottomrule
	\end{tabular}
	%}
	\label{table:computation_time}
\end{table}

Table \ref{table:computation_time} presents for our method and Micro-GPS the localization time,
without the required time for feature extraction.
The computational effort for feature extraction is comparable for both methods,
as it is dominated by the use of SIFT.
Using SiftGPU, feature extraction takes us about $40$\,ms.
The computational effort of our matching method grows linearly with the number of considered reference images;
for large numbers, it is slower than ANN matching approaches.
Accordingly, Micro-GPS performs global localization faster than our method.
However, in practice global localization is typically performed only once.
Afterwards, the previous pose estimation can be used as prior for the next localization step.
With increasing accuracy of the available prior,
less reference images have to be considered, reducing the localization time of our method.
If the prior is reliably more accurate than $1.5$\,m,
our method will be faster than Micro-GPS.
At the same time, as seen in Figure \ref{fig:with_prior},
the chance of correct localization increases when using a prior.

The memory consumption of our method is about three and a half times as large as that of Micro-GPS.
We roughly estimate the memory requirements as follows.
Per reference image,
Micro-GPS stores $50$ keypoints (with position, scale, and orientation) and $50$ $16$-dimensional floating point
descriptors,
resulting in $(50\cdot4\cdot32+50\cdot16\cdot32)$ bit $= 32000$ bit.
Our method stores per reference image $850$ keypoints (with position and orientation),
and a dictionary with $850$ pairs of $15$-bit descriptor values and integer feature indexes,
resulting in $(850\cdot3\cdot32+850\cdot(15+16))$ bit $=107950$ bit.

\section{CONCLUSION}
\label{sec:conclusion}
We examined methods for ground texture based absolute localization with and without available localization prior.
We propose identity matching, a feature matching strategy based on compact binary feature descriptors,
which simplifies feature matching to a single table lookup.
Substituting Micro-GPS's~\cite{Zhang_High-Prec-Localization} use of a global search index for feature matching with our strategy,
allowed us to reach higher localization success rates than the state-of-the-art methods for global localization.
Furthermore, our method allows to add, remove, and update mapped reference images online without the need of map recomputation.
Also, with our matching strategy the method is able to take advantage of prior pose estimates to perform local localization updates.
Apart from wooden floor texture, our method performs similarly well as state-of-the-art local localization methods,
while being faster to compute, especially for inaccurate prior pose estimates.
Lower computational cost can lead to higher effective localization accuracy,
as the time between image recoding and available pose estimation is shorter,
and it enables more frequent pose updates or savings on the required computational power.

In future research, we want to examine possible alternatives to the use of SIFT keypoints,
which make hardware acceleration necessary to reach reasonable computation times,
and we want to evaluate modifications to improve the localization capability on wooden texture.

%%%%%%%%%%%%%%%%%%%%%%%%%%%%%%%%%%%%%%%%%%%%%%%%%%%%%%%%%%%%%%%%%%%%%%%%%%%%%%%%
\addtolength{\textheight}{-12.5cm}
\bibliographystyle{IEEEtran}
\bibliography{../bib}

\end{document}